%% file: main.tex
\documentclass[letterpaper, 10 pt, conference]{ieeeconf}  % Comment this line out if you need a4paper
\usepackage[T1]{fontenc}
\usepackage{algorithm}% http://ctan.org/pkg/algorithms
\usepackage{xcolor}
\usepackage{algpseudocode}% http://ctan.org/pkg/algorithmicx
\usepackage{upgreek}
\usepackage{multicol}
\usepackage{ dsfont,setspace }
\usepackage{amsmath,graphicx}
\usepackage{tikz}
\usepackage{amssymb}
\usepackage{amsmath}

\usepackage{lineno}
\usepackage{epstopdf}
\usepackage{authblk}
\usepackage{dblfloatfix}
\usepackage{tabularx, makecell}
\newcolumntype{L}{>{\raggedright\arraybackslash}X}
\usepackage[
  separate-uncertainty = true,
  multi-part-units = repeat
]{siunitx}

\usepackage{subcaption}
\usepackage[colorlinks]{hyperref}
\hypersetup{
     colorlinks,
     linkcolor=black,
     citecolor=black,
     filecolor=black,
     urlcolor=blue,
 }
\usepackage{cleveref}
\usepackage{textcomp}
\usepackage{gensymb}

\IEEEoverridecommandlockouts                              % This command is only needed if 
\usepackage{amsmath} % assumes amsmath package installed
\usepackage{amssymb}  % assumes amsmath package installed

\begin{document}

\newcommand{\circled}[1]{\textcircled{\scriptsize #1}}
\newcommand{\circledSmall}[1]{\raisebox{0pt}{\textcircled{\raisebox{0.6pt}{\tiny #1}}}}

\title{\LARGE \bf
Comparison between External and Internal Single Stage Planetary gearbox actuators for legged robots
}

\author{
Aman Singh$^{1}$, Deepak Kapa$^{2}$, Prasham Chedda$^{3}$, Shishir N.Y. Kolathaya$^{1}$
\thanks{$^{1}$Aman Singh and Shishir N.Y. Kolathaya are with the Indian Institute of Science, Bengaluru, Karnataka, India.}
\thanks{$^{2}$Deepak Kapa is with the Indian Institute of Technology Roorkee, Roorkee, Uttarakhand, India.}%\
\thanks{$^{3}$Prasham Chedda is with the National Institute of Technology Tiruchirappalli, Tiruchirappalli, Tamil Nadu, India.}

}

\maketitle
\thispagestyle{empty}
\pagestyle{empty}

%%%%%%%%%%%%%%%%%%%%%%%%%%%%%%%%%%%%%%%%%%%%%%%%%%%%%%%%%%%%%%%%%%%%%%%%%%%%%%%%
\begin{abstract}

Legged robots, such as quadrupeds and humanoids, require high-performance actuators for efficient locomotion. Quasi-Direct-Drive (QDD) actuators with single-stage planetary gearboxes offer low inertia, high efficiency, and transparency. Among planetary gearbox architectures, Internal (ISSPG) and External Single-Stage Planetary Gearbox (ESSPG) are the two predominant designs. While ISSPG is often preferred for its compactness and high torque density at certain gear ratios, no objective comparison between the two architectures exists. Additionally, existing designs rely on heuristics rather than systematic optimization. This paper presents a design framework for optimally selecting actuator parameters based on given performance requirements and motor specifications. Using this framework, we generate and analyze various optimized gearbox designs for both architectures. Our results demonstrate that for the T-motor U12, ISSPG is the superior choice within the lower gear ratio range of 5:1 to 7:1, offering a lighter design. However, for gear ratios exceeding 7:1, ISSPG becomes infeasible, making ESSPG the better option in the 7:1 to 11:1 range. To validate our approach, we designed and optimized two actuators for manufacturing: an ISSPG with a 6.0:1 gear ratio and an ESSPG with a 7.2:1 gear ratio. Their respective masses closely align with our optimization model predictions, confirming the effectiveness of our methodology.
\end{abstract}


%%%%%%%%%%%%%%%%%%%%%%%%%%%%%%%%%%%%%%%%%%%%%%%%%%%%%%%%%%%%%%%%%%%%%%%%%%%%%%%%
\input{sections/introduction}
\input{sections/preliminaries}
\input{sections/methodology}
\input{sections/results}

\input{sections/conclusion}
\end{document}

%% file: sections/introduction.tex
\section{INTRODUCTION}

%% Motivation
% Why are legged robots useful?
Legged robots are transforming industrial and field robotics with advanced mobility in complex environments. Quadrupeds like Spot \cite{Spot} and ANYmal \cite{ANYmal} are used for facility inspection and construction surveying. Meanwhile, companies such as Tesla \cite{Tesla}, Agility Robotics \cite{AgilityRobotics}, and Unitree \cite{UnitreeHumanoid} are developing humanoid robots for general-purpose automation, further expanding robotic applications in industry. 

% QDD actuator
Many quadruped and humanoid robots use Harmonic drives for their high gear ratios and zero backlash, but their low efficiency necessitates torque sensors at the actuator output, increasing system complexity and cost. An alternative approach is quasi-direct-drive (QDD)\cite{CheetahActuator} actuators, which combine high torque density motors with low gear ratios to achieve high torque transparency and bandwidth. These designs typically employ single-stage planetary gearboxes to preserve these characteristics.
% Many quadruped and humanoid robot joints are powered by quasi-direct-drive (QDD) actuators \cite{CheetahActuator}, which utilise high-torque-density motors and low gear ratios to achieve high torque transparency and bandwidth. To maintain these characteristics, single-stage planetary gearboxes are commonly used in their design. 
There are two predominant architectures for such gearboxes: Internal Single-Stage Planetary Gearbox (ISSPG) \cite{MiniCheetah} and External Single-Stage Planetary Gearbox (ESSPG) \cite{MitCheetah3}. In ESSPG, the planetary gearbox is external to the motor’s body, making the design relatively straightforward. In contrast, ISSPG integrates the planetary gearbox inside the stator of the outrunner BLDC motor, resulting in a more compact and lightweight actuator. While ISSPG is generally favoured for its reduced size and higher torque density, there has been no objective study comparing the two architectures. Furthermore, the design of QDD actuators has largely relied on heuristics rather than systematic optimization \cite{MiniCheetah}\cite{MitCheetah3}.

Several studies \cite{PantherLeg, KaistHound} optimize actuator design parameters but do not consider overall mass or efficiency. Co-design approaches optimize both design and control, but works like \cite{StarlETHCooptimization, Co-designing_versatile_quadruped_robots_for_dynamic_and_energy-efficient_motions, Vitruvio, MetaRLCodesign} focus on link lengths, knee transmission ratios, and spring stiffness while overlooking gearbox design. Similarly, \cite{JointOptIFT} optimizes link lengths and actuator attachment points without addressing gearbox parameters.
Other studies \cite{StochasticProgCodesign1, StochasticProgCodesign2} apply co-design to manipulators and monopeds, optimizing gear ratios, compliance, link lengths, and mass but without evaluating gearbox type, efficiency, or mass. Additionally, \cite{A_versatile_co-design___legged_robots, Computational_design_of____size_and_actuators, Simulation_Aided_Co-Design_for_Robust_Robot_Optimization} model motor and gearbox friction but focus on belt drives rather than planetary gears.

In this paper, we introduce a methodology for optimizing the parameters of a single-stage planetary gearbox and, consequently, the actuator design, given specific joint performance requirements and a predefined motor. For a given joint performance requirement and motor, the required gear ratio of the gearbox can be determined. The gearbox is designed to adhere to this calculated gear ratio. Subsequently, the gearbox and actuator design are optimized to achieve minimal mass and maximum efficiency.

The efficiency model used in this work is based on \cite{BasicDrivingEff}, while the mass model incorporates not only the mass of the motor and gears but also all additional components necessary for a fully functional actuator. These include bearings, couplings, the carrier, the outer cover of the actuator, and other structural elements.
We optimize the designs of two different planetary gearbox architectures: the External Single-Stage Planetary Gearbox (ESSPG) and the Internal Single-Stage Planetary Gearbox (ISSPG). To provide a holistic understanding of the performance of these gearboxes, we analyse actuators operating at low gear ratios in the range of 5:1 to 15:1, which is a typical range for Quasi-Direct Drive (QDD) actuators. For each sub-range (e.g., 5:1–6:1, 6:1–7:1, ..., 14:1–15:1), we determine an optimal actuator for both gearbox architectures.

A comparative analysis of the actuators is performed by evaluating the cost functions associated with each design, which depend on mass and efficiency. This allows us to objectively determine which gearbox architecture—ISSPG or ESSPG—is preferable for a given requirement.
We have created an automated design framework to generate the parametric template model of the actuator. Following the optimization of the gearbox parameters, we use this automated design framework to generate preliminary design that can be refined by a human designer for actual manufacturing. 

The key contributions of this work are as follows:

\begin{itemize}
\item We propose a design framework for optimal actuator selection based on performance requirements and motor specifications, minimising mass and maximising efficiency with a detailed mass model for improved optimization over previous research.

\item We use this design framework to compare ISSPG and ESSPG in legged robotics. To the best of our knowledge, this is the first quantitative comparison of these two designs.

\item We validate our approach by designing ISSPG and ESSPG based actuators for manufacturing and comparing their masses with parametric model predictions.
\end{itemize}

The paper is organized as follows: Section \ref{sec:Preliminaries} covers the preliminaries of gearbox design, mass modeling, and efficiency modeling. Section \ref{sec:Methodology} details the methodology used to optimize actuators. Section \ref{sec:Results} presents simulation results and the design of manufacturable actuators. Finally, Section \ref{sec:Conclusion} concludes the paper and discusses future work.

%%%%% Harmonic drive

% 1. Harmonic drives have been used as legged robot actuators, but they are generally used for high gear ratios. In this paper, we wanted to analyze only single stage planetary gearboxes with low gear ratios (5:1- 15:1).

% 2. Harmonic drive provide zero backlash but as a trade-off their efficiency goes down. To avoid this problem in legged robot context, harmonic drives are generally used with a torque sensor at the actuator output. This complicates the design as well as makes the actuator expensive. 

% 3. Harmonic drive due to it's high torque capacity in limited volume and no backlash design is an important gearbox to analyze and we will be analyze it in the future work. 

% Many quadruped and humanoid robot joints are powered by Harmonic drives which have an advantage of high gear ratios, and zero backlash but suffer from reduced efficiency. To address the efficiency problem they are paired with torque sensors at the actuator output, increasing complexity and cost. Their is a different design philosophy that these legged robots use, which is called quasi-direct-drive(qdd) actuators, which utilize high torque density motors and low gear ratios to achieve high torque transparency and bandwidth. To maintain these characteristics, single-stage planetary gearboxes are commonly used in their design. 

%% file: sections/preliminaries.tex
\section{Preliminaries} \label{sec:Preliminaries}
% \subsection{Gearbox Types}
\subsection{Gearbox Designs}

% General introduction to the section

\begin{figure}[htbp]
    \centering
    \includegraphics[width=0.46\textwidth]{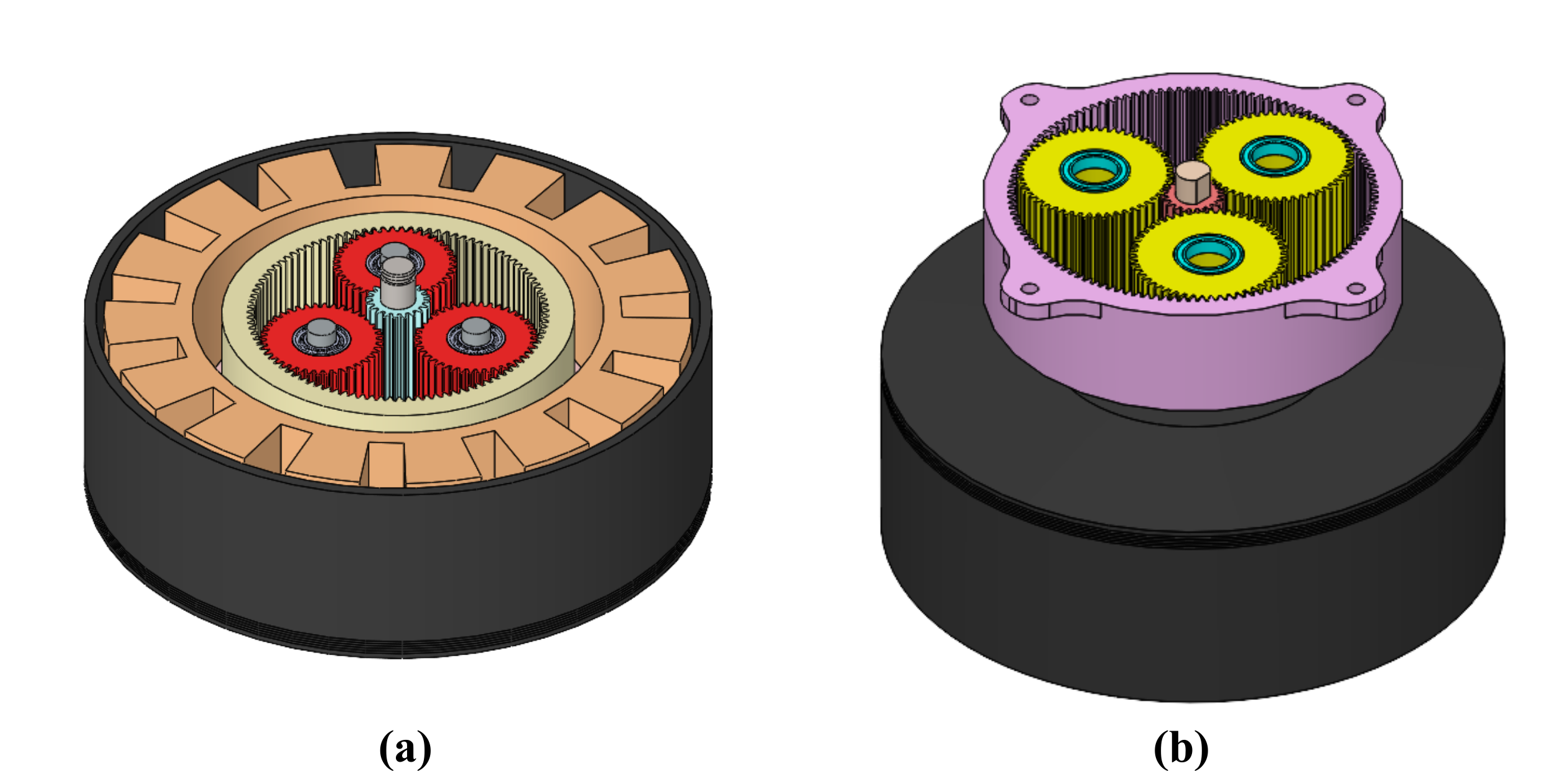}
    \caption {\textbf{(a) ISSPG:} Gearbox inside stator (orange); \textbf{(b) ESSPG:} Gearbox outside motor body (black).}
    \label{fig:ISSPG and ESSPG}
\end{figure}

This section describes Internal (ISSPG) and External (ESSPG) Single stage planetary gearboxes. The designs can be seen in the Fig. \ref{fig:ISSPG and ESSPG}

%-------------------------------------
% Single stage planetary Gearbox
%-------------------------------------
\subsubsection{External Single Stage Planetary Gearbox (ESSPG)}

The external single-stage planetary gearbox (ESSPG) is commonly used in legged robot actuators, including the MIT Cheetah 3~\cite{MitCheetah3}. We analyze a configuration with a fixed ring gear and motor-driven sun gear, which maximizes reduction. Single-stage designs are preferred for their high efficiency, since additional stages reduce it.

The gear reduction ratio and efficiency of a single-stage planetary gearbox are defined as:

\begin{equation}\label{Eq1}
G_{sspg} = \frac{\omega_c}{\omega_s} = \frac{N_s}{N_s + N_r}
\end{equation}

\begin{equation}\label{eq:eff_sspg_isspg}
\eta_{sspg} = \frac{N_s + \eta_{sp}\eta_{pr}N_r}{N_s + N_r}
\end{equation}

Here, $G_{sspg}$ is the gear ratio, where $\omega_c$ and $\omega_s$ are the carrier and sun gear angular velocities, and $N_s$, $N_r$ are the sun and ring gear teeth. $\eta_{sspg}$ is the overall efficiency, with $\eta_{sp}$ and $\eta_{pr}$ denoting sun-planet and planet-ring efficiencies.

%-------------------------------------
% Internal Single stage planetary Gearbox
%-------------------------------------
\subsubsection{Internal Single Stage Planetary Gearbox (ISSPG)}

The internal single-stage planetary gearbox is a compact, stator-integrated design, first introduced in the MIT Mini-Cheetah~\cite{MiniCheetah}. It leverages motor space to reduce actuator thickness and weight, increasing torque density. Note that it shares the same efficiency expression and reduction ratio as external single-stage planetary gearboxes.

\input{sections/mass_model}

\subsection{Efficiency Model}

The driving efficiency of the considered gearboxes depends on both gear tooth counts and the basic driving efficiencies of gear pairs. The sun–planet and planet–ring gear efficiencies are denoted by $\eta_a$ and $\eta_b$, respectively. This analysis follows the methodology described in~\cite{BasicDrivingEff}.

The basic driving efficiency formula is as follows:

% 1a. Basic Driving Efficiency formula
\begin{equation}\label{Eq_basic_dr_eff}
     \eta_i = 1 - \mu \pi \left( \frac{1}{N_{i1}} + sgn_i \frac{1}{N_{i2}}\right) \epsilon_{i}, i \in \{a, b\}
\end{equation}

% 1b. Explaining the symbols
where $\mu$ denotes the average friction coefficient of the tooth surface, the subscripts are defined by $a1 = s$, $a2 = p$, $b1 = p$, and $b2 = r$ for ESSPG and ISSPG. 

The sign parameters $sgn_i = +1$ if both the gears in the pair $i$ are external spur gears, if any one of the gears in the pair $i$ is internal gear then $sgn_i = -1$. So, $sgn_a = +1$ and $sgn_b = -1$. The parameters $\epsilon_{i}$ for $ i \in \{a, b\} $ are given by:
%----------------------------------------------
% 2. Formulas for contact ratio
\begin{equation}\label{eps_param}
     \epsilon_i = \epsilon_{i1}^2 + \epsilon_{i2}^2 - \epsilon_{i1} - \epsilon_{i2} + 1
\end{equation}

% 2a. Approach contact ratio
\begin{equation}\label{Approach_CR}
     \epsilon_{i1} = sgn_{i} \frac{N_{i2}}{2 \pi} \left( tan(\alpha_{a_{i2}}) - tan(\alpha) \right)
\end{equation}

% 2b. Recess contact ratio
\begin{equation}\label{Recess_CR}
     \epsilon_{i2} = \frac{N_{i1}}{2 \pi} \left( tan(\alpha_{a_{i1}}) - tan(\alpha) \right)
\end{equation}

where $\epsilon_{i1}$ and $\epsilon_{i2}$ are the approach and recess contact ratio, respectively, and $\alpha$ is the basic pressure angle. Also, 
%$\alpha_{w_i}$ is the working pressure angle, and 
$\alpha_{a_{i1}}$ and $\alpha_{a_{i2}}$ are the tip pressure angles. 
Please note the above expressions of $\epsilon_{i1}$ and $\epsilon_{i2}$ are true only when the profile shift coefficients in all the gears are $0$. We have assumed them to be $0$ in our current analysis. 

The tip pressure angle is given by:

% 3b. Tip Pressure Angle
\begin{equation}\label{Eq_basic_dr_eff}
     \cos \alpha_{a_j} = \frac{d_{b_j}}{d_{a_j}}
\end{equation}

where, $j \in \{s, p ,r \}$ for both ESSPG and ISSPG.
% $j \in \{s, p1, p2, r \}$ for CPG. 
Here, $d_{b_j}$ represents the basic circle diameter and is given by $d_{b_j} = m N_{j} \cos(\alpha)$ where $m$ represents the module of gears. Also, $d_{a_j}$ represents the tip circle diameters of the gears. The values of the tip circle diameters of ESSPG and ISSPG gearboxes are given as follows:

%----------------------------------------------

% 4a. Tip Circle Diameters
\begin{equation}\label{Eq_basic_dr_eff}
     d_{a_j} = m N_j +  sgn_{j} \cdot 2m, \ j \in \{s, p, r\}
\end{equation}
% \begin{equation}\label{Eq_basic_dr_eff}
%      d_{a_p} = m N_p + 2 m
% \end{equation}
% \begin{equation}\label{Eq_basic_dr_eff}
%      d_{a_r} = m N_r - 2 m
% \end{equation}
where $d_{a_s}, d_{a_s},$ and $d_{a_r}$ represents the tip circle diameters of the sun, the planet, and the ring gear, respectively, $sgn_{j} = +1$ if $j \in \{s,p\}$ and $sgn_r = -1$. 

% 7. The independent variables 
Here, number of teeth, module, and pressure angle are free variables. The pressure angle is taken to be $\alpha = 20 ^\circ$.

%----------------------------------------------
\subsection{Lewis Model for Gear Strength}

The Lewis Bending Equation is used to determine the face width of gears based on gear parameters and material properties. To account for dynamic effects, a modified version of the Lewis Bending Equation is employed by incorporating the velocity factor ($K_v$). The modified equation is expressed as:  

\begin{equation}
     \sigma = \frac{F_t}{b \cdot y \cdot K_v \cdot P}
\end{equation}  

Where, $\sigma$ is the bending stress, $K_v$ is the velocity factor, $F_t$ is the tangential component of the gear tooth, $P$ is the circular pitch, $b$ is the face width, and $y$ is the Lewis Form Factor. The formula of the Lewis form factor and velocity factor taken from\cite{TKrishnaRao}.

Where, $V_m$ is the velocity $V_m = \omega \cdot r$ and $N$ is the number of teeth.

To express the face width as a function of gear parameters and bending stress, the equation is rearranged as:  

\begin{equation}
     b = \frac{\text{FOS} \cdot F_t}{\sigma \cdot y \cdot K_v \cdot P}
\end{equation}  

where FOS is the factor of safety. 

The optimization algorithm utilizes this equation to compute the face width, which is subsequently used to determine the gear volume and mass.

%% file: sections/mass_model.tex
\subsection{Mass Model} \label{sec:MassModel}

A detailed mass model is essential to make precise comparisons between different design architectures. Simply comparing gear masses is inadequate, as gearbox type affects overall actuator structure and mass.

\subsubsection{Parametric Mass Model Assumptions and Design}

The parametric mass model takes as input motor dimensions, gearbox type, and the number of sun, planet, and ring teeth. Actuator dimensions are motor-constrained. Gear face widths are computed using the Lewis strength model [Section~2.4]. Using these inputs, the dimensions and masses of components—main casing, base plate, and bearings—are then calculated.

Spur gears are modeled as cylinders with pitch-circle diameters and central bores. Internal gears, such as ring gears, include added radial thickness. Bearings are chosen from a continuous regression model derived from datasheets to optimize actuator mass, maintain performance despite discrete sizing, and ensure continuity of the mass function.

Carriers and secondary carriers are modeled as hollow disks with symmetric planet and support pins. Their dimensions depend on bearing size, number of planets, and gear size. The casing is a hollow cylinder constrained by the motor’s outer diameter, with thickness varying by local load. Small components (e.g., nuts, bolts, circlips) are excluded due to negligible mass, offset by material at bores. Gears use cast carbon steel; other parts use aluminum.

%% file: sections/methodology.tex
\begin{figure*}[htbp]
    \centering
    \includegraphics[width=\textwidth]{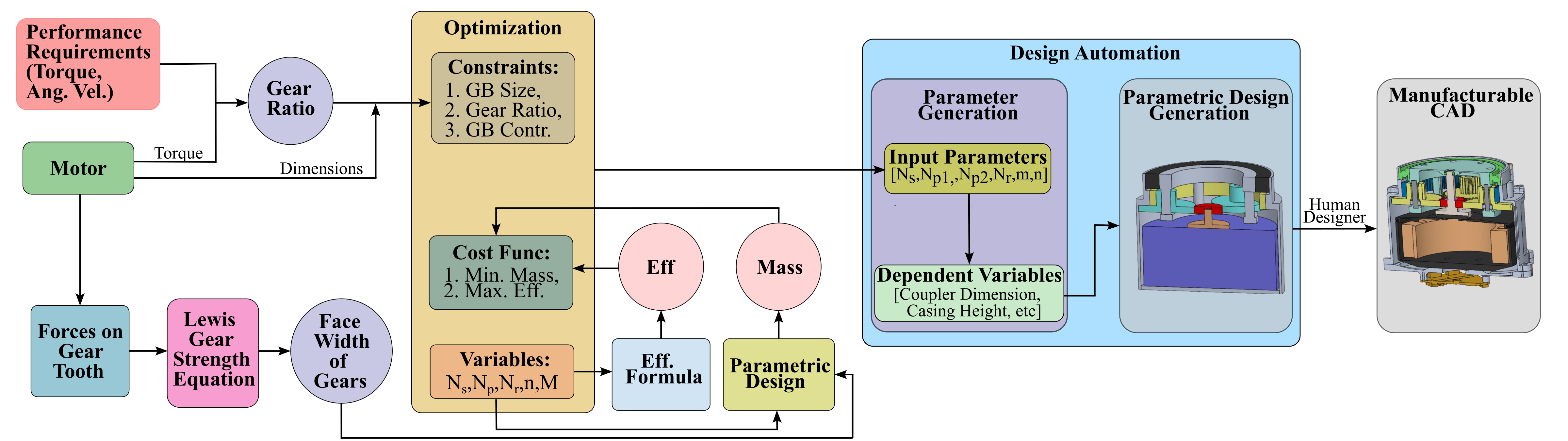} % Adjust the width as needed
    \caption{\textbf{Methodology:} The optimization framework (left) optimizes gearbox parameters for a given motor and performance requirements, passing them to the design automation block. This generates a parametric template model, which a human designer uses to create the manufacturable CAD of the actuator.}
    \label{fig:flowchart2}
\end{figure*}

\section{Methodology}\label{sec:Methodology}
\subsection{Optimization of Design}

This section formulates the actuator gearbox optimization problem. Subsequent subsections define the variables for different gearbox configurations, outline constraints for feasible and efficient designs, and describe cost functions used to obtain optimal actuator designs. The overall optimization methodology is shown in Fig.~\ref{fig:flowchart2}.

\subsubsection{Variables}
Although many parameters influence gearbox performance, we focus on variables with significant impact on optimization. For both ESSPG and ISSPG—being single-stage planetary gearboxes—the design variables are identical and defined as:
\begin{equation}\label{Opt_var}
    X := [ N_{s}, N_{p}, N_{r}, m, n_p]
\end{equation}
Here, $N_{s}, N_{p}, N_{r}$ are the number of teeth on the sun, planet, and ring gears; $m$ is the gear module; and $n_p$ is the number of planet gears. $N_{s}, N_{p}, N_{r}, n_p$ are integer variables, and $m$ is a discrete variable.

% Constraints
\subsubsection{Constraints:}

\begin{figure}[H]
    \centering
    \includegraphics[width=0.48\textwidth]{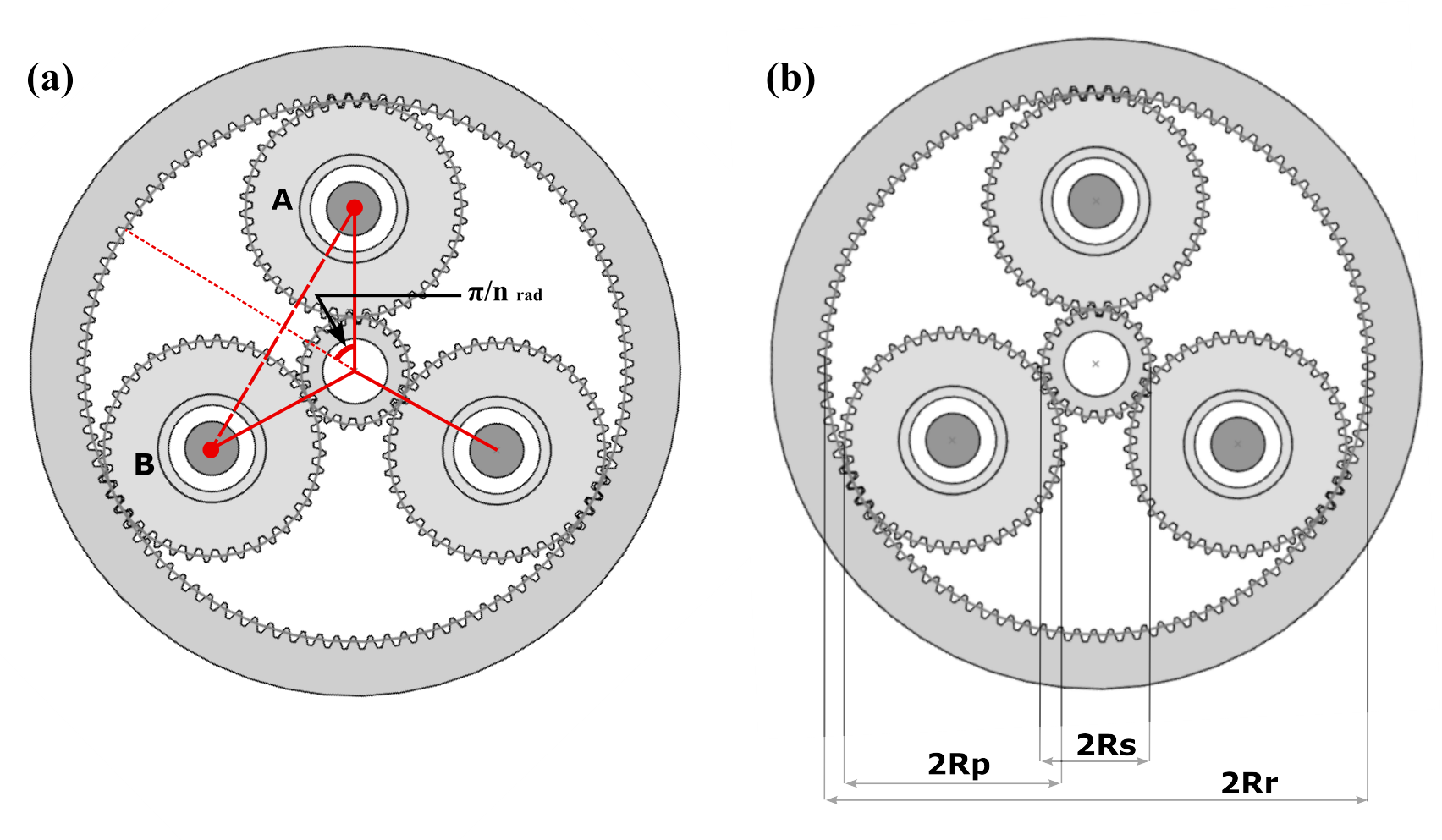}
    \caption{Meshing and Geometric Constarints}
    \label{fig:Meshing and Geometric Constraint}
\end{figure}

There are several constraints used in the optimization problem formulation. The description of each constraint is given as follows:

The \textbf{geometric constraint} for both ESSPG and ISSPG ensures dimensional compatibility and is defined as (Fig.~\ref{fig:Meshing and Geometric Constraint}):

\begin{equation}\label{eq:geo_constr_sspg_isspg}
    N_r = N_s + 2N_p 
\end{equation}

%%%%%%%% Figure 

The \textbf{meshing constraint} ensures proper engagement between contacting gears. For SSPG and ISSPG, the sum of sun and ring gear teeth must be divisible by the number of planet gears~\cite{sspgTeethMatchingCond}, expressed as:

\begin{equation}\label{eq:mesh_constr_sspg_isspg}
N_s + N_r = n_p a,\quad a \in \mathbb{Z}^{+}
\end{equation}
where a is a positive integer.

The \textbf{No interference constraint} ensures planet gears do not overlap, limiting the number that can be accommodated. For ESSPG and ISSPG, it is defined as:
\begin{equation}\label{eq:no_planet_interference_constr_sspg_isspg}
    2m (N_s + N_p) \sin\left(\frac{\pi}{n_p}\right) - 2mN_p \geq \delta_{p}
\end{equation}
where $\delta_{p} = 5$ mm is the minimum clearance between planet gears.

Several \textbf{other constraints} limit gear module, tooth count, and number of planet gears for ESSPG and ISSPG:

\begin{equation}\label{eq:other_constr}
    \begin{split}
    & \ m_{min} \leq m \leq m_{max} \\
    & \ N_{min} \leq N_s, N_{p} \\
    % & \ m N_s \leq D^{max}_{GB} \\
    % & \ m N_p \leq D^{max}_{GB} \cdot 0.5 \\
    & \ m N_r \leq D^{max}_{GB} \\
    & n^{min}_p \leq n_p \leq n^{max}_p
    \end{split}
\end{equation}

The gear module is constrained between $m_{\min} = 0.5$ mm and $m_{\max} = 1.2$ mm, as per manufacturing limits provided by the supplier. To avoid undercutting, the minimum number of teeth is set to $N_{\min} = 20$ for both sun ($N_s$) and planet ($N_p$) gears. While there is no explicit upper bound on the number of teeth, they are implicitly limited by the maximum allowable gearbox diameter, $D^{\max}_{GB}$. For ESSPG, $D^{\max}_{GB} = D^{\text{motor}}_{OD} - \delta_{\text{clr}}$, and for ISSPG, $D^{\max}_{GB} = D^{\text{motor}}_{ID} - \delta_{\text{clr}}$, where $D^{\text{motor}}_{OD}$ and $D^{\text{motor}}_{ID}$ are the motor’s outer and stator internal diameters, respectively, and $\delta_{\text{clr}} = 10$ mm is the ring gear clearance. The number of planet gears, $n_p$, is restricted to the range $[2, 7]$, in accordance with values commonly reported in literature.

\subsubsection{Optimization Problem}

The objective is to determine an optimal gearbox design for a given motor by minimizing actuator mass and maximizing efficiency. The cost function is defined as
\begin{equation}\label{eq:cost}
    \text{cost} := K_m M_{act} - K_e \eta,
\end{equation}
where $M_{act}$ is actuator mass, $\eta$ is efficiency, and $K_m = 1$, $K_e = 2$ are user-defined constants. Adjusting these constants allows prioritizing mass or efficiency (e.g., $K_m \gg K_e$ favors mass). Here, efficiency is slightly prioritized. Note that $0 \leq \eta \leq 1$. Efficiency and mass are computed using Eq.~\ref{eq:eff_sspg_isspg} and Sec.~\ref{sec:MassModel}, respectively.

A brute-force search is used to solve the optimization problem by evaluating all feasible points across gear ratios, computing the cost function, and selecting the design with minimum cost. This process is repeated for all design types to determine the optimal design and corresponding parameters for each gearbox range.

\subsection{Automation of Design}

We developed a framework to generate 3D CAD models of parametric actuators. It enables visualization and modification of the mass model to align with real-world actuator characteristics, aiding in validation and verification. The framework also serves as a template for designing manufacturable actuators. Users can input key parameters, including motor dimensions (inner diameter, outer diameter, height) and gearbox specifications: number of teeth on sun, planet, and ring gears ($N_{\text{s}}, N_{\text{p}}, N_{\text{r}}$), gear module ($m$), and number of planet gears ($n_p$). Supplementing these with additional design variables, the program computes dependent variables—such as outer casing diameter, height, radial width, and carrier dimensions—validates inputs, and exports them to CAD software to generate the 3D model.

%% file: sections/results.tex
\input{sections/results_table_SSPG}

\input{sections/results_table_iSSPG}

\section{Results}\label{sec:Results}
This section presents simulation results from the optimization problem and parametric designs generated through the automated design framework. We then discuss actuator designs for manufacturing, created with human input aided by automated parametric design. We also assess feasibility and compare actuator weights with the automated design predictions.

The motor used for optimization and design is the T-motor U12 \cite{TmotorU12}, an outrunner BLDC motor with high torque density and a large gap radius. While the U12 motor was used for analysis, the proposed methodology is generalized and can be easily extended to other motors.

\begin{figure}[htbp]
    % \centering
    \includegraphics[width=\linewidth]{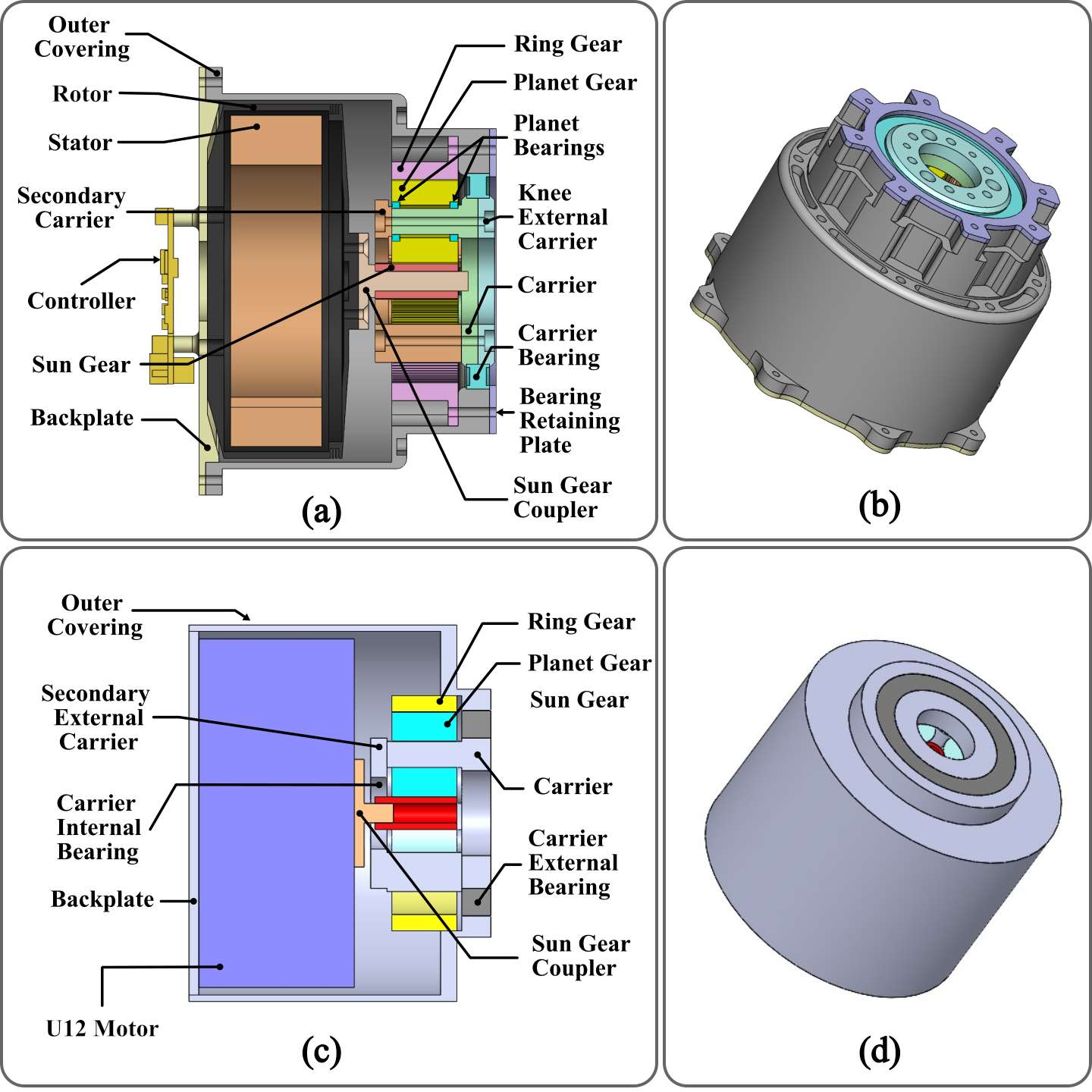}
    \caption{ESSPG Actuator with gear ratio 7.2:1; (a, b) Design for Manufacturing; (c, d) Template Designs}
    \label{fig:sspg}
\end{figure}

\begin{figure}[htbp]
    % \centering
    \includegraphics[width=\linewidth]{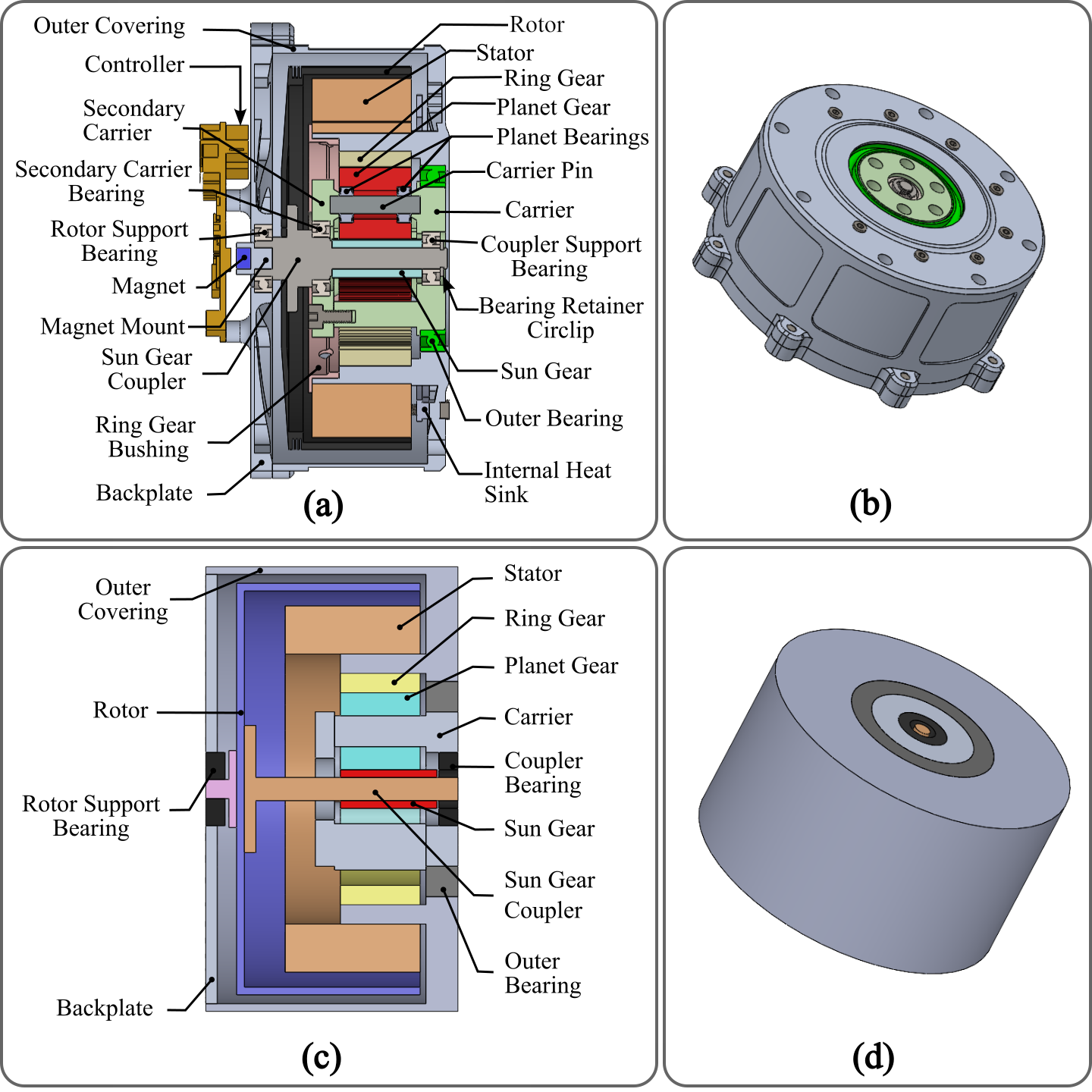}
    \caption{ISSPG Actuator with gear ratio 6:1; (a, b) Design for Manufacturing; (c, d) Template Designs}
    \label{fig:isspg}
\end{figure}

\subsection{Simulation Results} 
% We solved the optimization problem for both Internal and external single stage planetary gearbox to determine the optimal design for each gear ratio range. Optimal actuators were identified for the two gearbox classes across gear ratios from 5 to 15, with steps of one (i.e., 5–6, 6–7, and so on). The resulting data is presented in tables \ref{table:sspg_results} and \ref{table:isspg_results}.

We solved the optimization problem for internal and external single-stage planetary gearboxes to identify optimal designs for gear ratios 5–15 (step size 1). The corresponding actuator configurations for each gearbox class are summarized in Tables~\ref{table:sspg_results} and~\ref{table:isspg_results}.

\begin{itemize}
    \item \textbf{Remark 1:} The internal single-stage planetary gearbox was lighter than the external single-stage planetary gearbox for the 5-7 gear ratio range. But the efficiency of the internal and external gearboxes was same. This made ISSPG a better choice than ESSPG, in this range.
    % \item \textbf{Remark 2:} The external singl planetary gearbox was the most optimal for the 7–15 gear ratio range.
    \item \textbf{Remark 2:} The internal single-stage planetary gearbox did not yield any actuators for the 7–15 gear ratio range, as the restriction on the maximum gearbox size dictated by the stator inner diameter prevented achieving gear ratios higher than 7.
    \item \textbf{Remark 3:} The external single-stage planetary gearbox did not yield any actuators for the 11–15 gear ratio range, as the restriction on the maximum gearbox size dictated by the motor size prevented achieving gear ratios higher than 11.

\end{itemize}

\subsection{Manufacturable Designs}

After solving the optimization problem for each gear ratio range, we utilized the automated design framework to generate parametric models of the optimal actuators of each of the two types. We selected gear ratios, 6.0:1 (optimal in 6–7 gear ratio range) for ISSPG and 7.2:1 (optimal in 7-8 gear ratio range) for ESSPG, and developed their manufacturable designs using the parametric templates as references, as illustrated in Fig. \ref{fig:sspg} and Fig. \ref{fig:isspg}. The detailed designs closely matched the parametric models in mass, with the ESSPG 7.2:1 actuator weighing 1.657 kg (model) and 1.663 kg (detailed), and the ISSPG 6.0:1 actuator weighing 1.386 kg (model) and 1.345 kg (detailed). %\ref{table:mass_comparison}.
A slight difference exists between the mass of the model design and that of the detailed design, which is difficult to quantify, as creating detailed actuator models for all gear ratios is impractical. This discrepancy arises from modeling bearing parameters as continuous functions of inner diameter, omitting holes in the model, and simplifying spur gears as solid cylinders.

% You can see there is a small difference between the mass of our model design and the mass of the detailed design. It is difficult to quantify this differnece because it is very difficult to make detailed design of an actuator for each and every gear ratio. But the possible reasons for this difference can be following:

% 1. We have modeled the bearing parameters (like mass, Outer diameter, width) as a continuous function of the inner diameter, even though it is a discrete function. 

% 2. In our model design there are no holes modeled to keep the model simple. 

% 3. Also, the spur gears have been modeled as cylinders to keep the model simplified. 

The actuators have not yet been manufactured, and their efficiency can only be validated through hardware testing. The reported masses were computed using CAD models with assigned material properties, and standard components such as bearings and motor drivers were weighted based on datasheet specifications.

%% file: sections/results_table_sspg.tex
\begin{table}[h]
\caption{ESSPG results}
\label{table:sspg_results}
\centering
\resizebox{0.95\linewidth}{!}{ % Adjust 0.9 to make it smaller/larger
\begin{tabular}{|c|c|c|c|c|c|}
    \hline
    \textbf{$G_r$} & \textbf{$(m^*, N_s, N_p, N_r, n)$} & \textbf{Mass} & \textbf{$\eta$(\%)} & \textbf{Cost} & \textbf{$\tau_d$} \\
     &  & (Kg) & & & (Nm) \\
    \hline
    5.0 & (0.6, 20, 30, 80, 4) & 1.375 & 90.9 & 1.84 & 32.11 \\
    \hline
    6.0 & (0.5, 20, 40, 100, 3) & 1.508 & 92.3 & 2.09 & 35.13 \\
    \hline
    7.2 & (0.5, 20, 52, 124, 3) & 1.657 & 93.3 & 2.38 & 38.37 \\
    \hline
    8.1 & (0.5, 20, 61, 142, 3) & 1.823 & 93.7 & 2.71 & 39.23 \\
    \hline
    9.0 & (0.5, 20, 70, 160, 3) & 1.959 & 94.1 & 2.98 & 40.57 \\
    \hline
    10.2 & (0.5, 20, 82, 184, 3) & 2.224 & 94.4 & 3.50 & 40.50 \\
    \hline
\end{tabular}
}
\begin{minipage}{\linewidth}
\footnotetext{*The module $m$ is expressed in millimeters (mm).}
\end{minipage}
\end{table}

%% file: sections/results_table_isspg.tex
\begin{table}[h] 
\caption{ISSPG Results}
\label{table:isspg_results}
\centering
\resizebox{0.95\linewidth}{!}{%
\begin{tabular}{|c|c|c|c|c|c|}
    \hline
    \textbf{$G_r$} & \textbf{$(m^*, N_s, N_p, N_r, n)$} & \textbf{Mass} & \textbf{$\eta$ (\%)} & \textbf{Cost} & \textbf{$\tau_{d}$} \\
     &  & (Kg) & & & (Nm) \\
    \hline
    5.0 & (0.6, 20, 30, 80, 4) & 1.265 & 90.9 & 1.62 & 34.913 \\
    \hline
    6.0 & (0.5, 20, 40, 100, 3) & 1.386 & 92.3 & 1.85 & 38.213 \\
    \hline
\end{tabular}
}
\begin{minipage}{\linewidth}
% \footnotetext{*The module $m$ is expressed in millimeters (mm).}
\end{minipage}
\end{table}

%% file: sections/conclusion.tex
\section{Conclusion} \label{sec:Conclusion}
This paper presented a systematic design framework for optimizing actuator parameters based on performance requirements and motor specifications. By applying this framework, we analyzed and compared optimized designs for both ISSPG and ESSPG architectures. Our results indicate that for the T-motor U12, ISSPG is preferable within the lower gear ratio range (5:1 to 7:1) due to its lighter design, whereas ESSPG becomes the superior choice for higher gear ratios (7:1 to 11:1). To validate our approach, we designed and optimized two actuators for manufacturing, demonstrating strong alignment between predicted and actual masses, confirming the effectiveness of our methodology.
Future work will focus on extending the optimization framework to other gearbox designs, including the different types of planetary gearboxes, harmonic drives, cycloidal drives, incorporating motor parameters into the optimization process, and manufacturing and experimentally validating the actuators developed in this study.

\bibliographystyle{IEEEtran}
\bibliography{references}